\documentclass[sigconf]{acmart}
\AtBeginDocument{%
  }

\setcopyright{acmlicensed}

\copyrightyear{2026}
\acmYear{2026}
\setcopyright{cc}
\setcctype{by}
\acmConference[MM '26]{Proceedings of the 34th ACM International Conference on Multimedia}{November 10--14, 2026}{Rio de Janeiro, Brazil}
\acmBooktitle{Proceedings of the 34th ACM International Conference on Multimedia (MM '26), November 10--14, 2026, Rio de Janeiro, Brazil}
\acmDOI{10.1145/3767308.3835163}
\acmISBN{979-8-4007-2213-4/2026/11}
\begin{document}

%%
%% The "title" command has an optional parameter,
%% allowing the author to define a "short title" to be used in page headers.
\title{DAS-PMVC: A Framework for Partial Multi-View Clustering via Dual Alignment and Structure Enhancement}

%%
%% The "author" command and its associated commands are used to define
%% the authors and their affiliations.
%% Of note is the shared affiliation of the first two authors, and the
%% "authornote" and "authornotemark" commands
%% used to denote shared contribution to the research.
\author{Shubin Ma}
\affiliation{%
  \institution{Dalian University of Technology}
  \city{Dalian}
  \state{Liaoning}
  \country{China}
}
% \authornote{Both authors contributed equally to this research.}
\email{shubinma@mail.dlut.edu.cn}
\orcid{0009-0002-9794-2661}

\author{Liang Zhao}
\correspondingauthor
% \authornotemark[1]
\orcid{0000-0001-6301-1311}
\affiliation{%
  \institution{Dalian University of Technology}
  \city{Dalian}
  \state{Liaoning}
  \country{China}
}
\email{liangzhao@dlut.edu.cn}

\author{Chuanye He}
\affiliation{%
  \institution{Dalian University of Technology}
  \city{Dalian}
  \state{Liaoning}
  \country{China}
}
\email{2838527538@mail.dlut.edu.cn}

\author{Zhenjiao Liu}
\affiliation{%
  \institution{Inspur Group Co., Ltd.}
  \city{Jinan}
  \state{Shandong}
  \country{China}
}
\email{liuzhenjiao@inspur.com}

\author{Liang Zou}
\affiliation{%
  \institution{China University of Mining and Technology}
  \city{Xuzhou}
  \state{Jiangsu}
  \country{China}
}
\email{liangzou@cumt.edu.cn}

\author{Lin Yuanbo Wu}
\affiliation{%
  \institution{The University of Warwick}
  \city{Coventry}
  \state{West Midlands}
  \country{United Kingdom}
}
\email{Yuanbo.Lin@warwick.ac.uk}

\author{Yu Shao}
\affiliation{%
  \institution{Dalian University of Technology}
  \city{Dalian}
  \state{Liaoning}
  \country{China}
}
\email{python@mail.dlut.edu.cn}

%%
%% By default, the full list of authors will be used in the page
%% headers. Often, this list is too long, and will overlap
%% other information printed in the page headers. This command allows
%% the author to define a more concise list
%% of authors' names for this purpose.
\renewcommand{\shortauthors}{Ma et al.}

%%
%% The abstract is a short summary of the work to be presented in the
%% article.
\begin{abstract}
  In recent years, multi-view clustering has attracted widespread research interest. However, due to limitations in data collection devices, data across different views often suffer from misalignment, leading to the partial view alignment problem (PVAP). To mitigate the impact of view asymmetry and irrelevant samples, this paper proposes a framework for partial multi-view clustering via dual alignment and structure enhancement (DAS-PMVC), which leverages view structure consistency and semantic relevance. Specifically, DAS-PMVC includes three parts: \textbf{anchor graph structure alignment}, where sample joint embedding representations with consistent latent space are derived from anchor point relationships for initial view alignment; \textbf{structure-enhanced feature learning}, where the model learns view structure information through pretraining and combines multi-view graph convolutional networks to further extract deep latent features from the aligned graph structure to improve the discriminative power of representations; and \textbf{a dual alignment strategy}, where initial alignment is performed through the anchor graph in the pretraining phase, and contrastive learning loss and the Hungarian algorithm are introduced in the training phase to further optimize the alignment of latent features. Experimental results on various datasets demonstrate that the DAS-PMVC framework outperforms existing state-of-the-art methods in clustering performance, showcasing its effectiveness and superiority.
\end{abstract}

%%
%% The code below is generated by the tool at http://dl.acm.org/ccs.cfm.
%% Please copy and paste the code instead of the example below.
%%


%%
%% Keywords. The author(s) should pick words that accurately describe
%% the work being presented. Separate the keywords with commas.
%% A "teaser" image appears between the author and affiliation
%% information and the body of the document, and typically spans the
%% page.

% \received{20 February 2007}
% \received[revised]{12 March 2009}
% \received[accepted]{5 June 2009}

%%
%% This command processes the author and affiliation and title
%% information and builds the first part of the formatted document.
\maketitle

\section{Introduction}
\label{sec1}
\begin{figure}[t]
	\centering
		\includegraphics[width=0.47\textwidth]{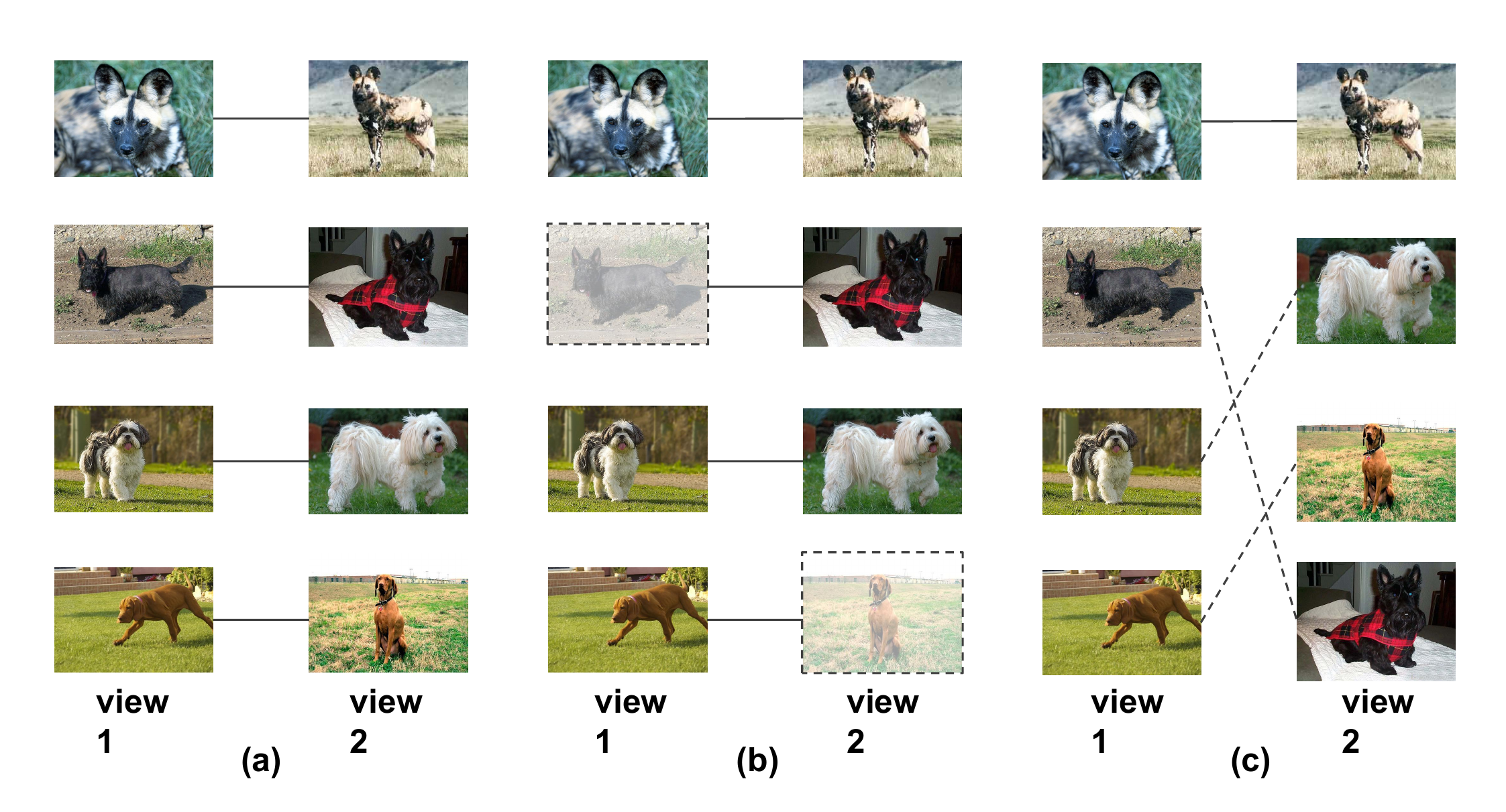}
	\caption{Example of multi-view data in different scenarios. Taking two views as an example, (a) indicates that the two views have a complete correspondence (alignment between samples and no missing samples). The dashed lines in (b) represent the missing samples in each view. The solid lines in (c) represent aligned samples, while the dashed lines represent samples that are not aligned.}
	\label{FIG:1}
\end{figure}

 In multi-view clustering \cite{kumar2011co,li2022high,zhao2025dynamic}, it is commonly assumed that there exists a complete correspondence between each view \cite{tao2018multiview,zhao2025dual}; however, this assumption is often difficult to achieve in practical scenarios \cite{zong2018multi,wang2020icmsc}. Hardware failures or noise during data collection can result in the loss of correspondence between different views of the same sample. Furthermore, multi-view data may be collected by different devices, and their processing and storage could be independent \cite{yu2021novel}. 
As a result, ensuring complete alignment of multi-view data is nearly impossible, leading to the PVAP.
Although converting partially aligned view samples into incomplete data by discarding misaligned data \cite{gong2022gromov} can be used for clustering analysis, this solution is not optimal. This is because transforming the data into incomplete multi-view data \cite{hu2019doubly,liu2018late} means that many samples only retain one available view, which significantly increases processing complexity when the number of views is large. Therefore, effectively handling such misaligned multi-view data presents a major challenge. Instead of treating it as incomplete multi-view data, it would be more prudent to find a more reasonable data realignment method.
Figure \protect\ref{FIG:1} illustrates the scenarios of incomplete views and partial view alignment in multi-view data.

For the more challenging problem of partially view-aligned multi-view clustering, several methods have been explored in recent years to address this issue. For instance, maximum covariance analysis (MCA) \cite{beauchaine2002comparison} can be performed on aligned data \cite{yang2021partially,ijcai2025ma}, followed by progressively obtaining correspondence on the optimal cost matrix using the Hungarian algorithm \cite{priya2019hungarian}. Alternatively, clustering each view based on non-negative matrix factorization (NMF) \cite{qian2016double,lin2025representation} and then using the clustering results and partial alignment information to establish correspondence for misaligned data has been proposed. These methods have major limitations. Firstly, these methods are shallow models; secondly, they establish view correspondence in separate steps. It is highly desirable to jointly perform view alignment and downstream tasks to leverage the representational power of neural networks. 

Specifically, some methods aim to cluster \cite{Liu2026info, Guan2025mul} on unmapped data. Yu et al. proposed a novel framework based on NMF, which utilizes the consistency of view geometry to establish correspondence between views \cite{yu2021novel}. Zhang et al. used known mapping relationships to form constraints and proposed a constrained multi-view clustering algorithm for unmapped data \cite{zhang2015constrained}. For partially aligned view clustering, Huang et al. proposed an insertable alignment module that can be used in neural networks \cite{huang2020partially}. During training, the neural network can gradually learn to align misaligned data. Yang et al. adopted a contrastive learning scheme where paired view data of aligned samples constitute positive samples and other misaligned data randomly form negative samples \cite{yang2021partially, zhang2020hard}. Additionally, a new noise-resistant contrastive loss was designed to mitigate the negative impact of incorrect negative samples. Fu et al. address the view-misaligned problem by introducing a reorder matrix and propose a reordered k-means method based on this matrix, relaxing the constraints of traditional k-means to significantly improve clustering performance on view-misaligned datasets \cite{fu2024reordered}. Gao et al. introduce a triple-consistency-driven representation fusion method, optimizing view alignment and clustering fusion by separately extracting graph structure consistency, semantic consistency, and cluster consistency, thereby improving clustering performance under partially view-aligned conditions \cite{gao2025novel}. However, the above methods: 1) fail to effectively leverage the structural information of the views, overlooking the structural consistency between views, and 2) perform alignment only once using the trained samples, resulting in substantial alignment errors.

To address the limitations of existing methods, this paper proposes a framework for partial multi-view clustering via dual alignment and structure enhancement(DAS-PMVC). DAS-PMVC employs a dual alignment strategy to recover the correspondences between different views, while leveraging structural information and feature propagation techniques from graph learning to improve clustering accuracy. Specifically, to achieve initial alignment between views and overcome alignment challenges caused by spatial heterogeneity, DAS-PMVC identifies anchor points within multi-view data and constructs anchor graphs for each view. To fully utilize the graph structure of multi-view data, the framework incorporates a pre-training strategy to extract deep latent features from the aligned graph structure, enhancing the discriminative power of the representations. Additionally, to reduce errors introduced by initial alignment and ensure structural consistency and semantic invariance across views, DAS-PMVC implements dual structural guidance throughout the process, performing secondary alignment on the samples. Moreover, contrastive learning is introduced to accelerate the alignment between aligned and misaligned samples, generating more discriminative features. 
The main contributions of this work can be summarized as follows,

\begin{enumerate}
\item \textbf{Anchor Graph Structure Alignment:} The anchor point relationships are used to derive joint embedding representations of samples with consistent latent space, constructing anchor graphs. Structural filtering is then applied to the anchor graph to effectively reduce the impact of irrelevant samples, achieving initial alignment across views.
\item \textbf{Structure-Enhanced Feature Learning:} The model learns view structure information through pre-training and utilizes multi-view graph convolutional networks to extract deep latent features from the aligned graph structure. Dual structural guidance is implemented to enhance the discriminative power of the learned representations.
\item \textbf{Dual Alignment Strategy:} The initial alignment is achieved through anchor graphs during the pre-training phase to enhance cross-view consistency. In the training phase, contrastive learning loss and the Hungarian algorithm are introduced to further optimize the alignment of latent features, reducing the errors from the initial alignment.

\end{enumerate}

\section{Related Work}

\begin{figure*}[ht]
\centering
\includegraphics[width=0.98\textwidth]{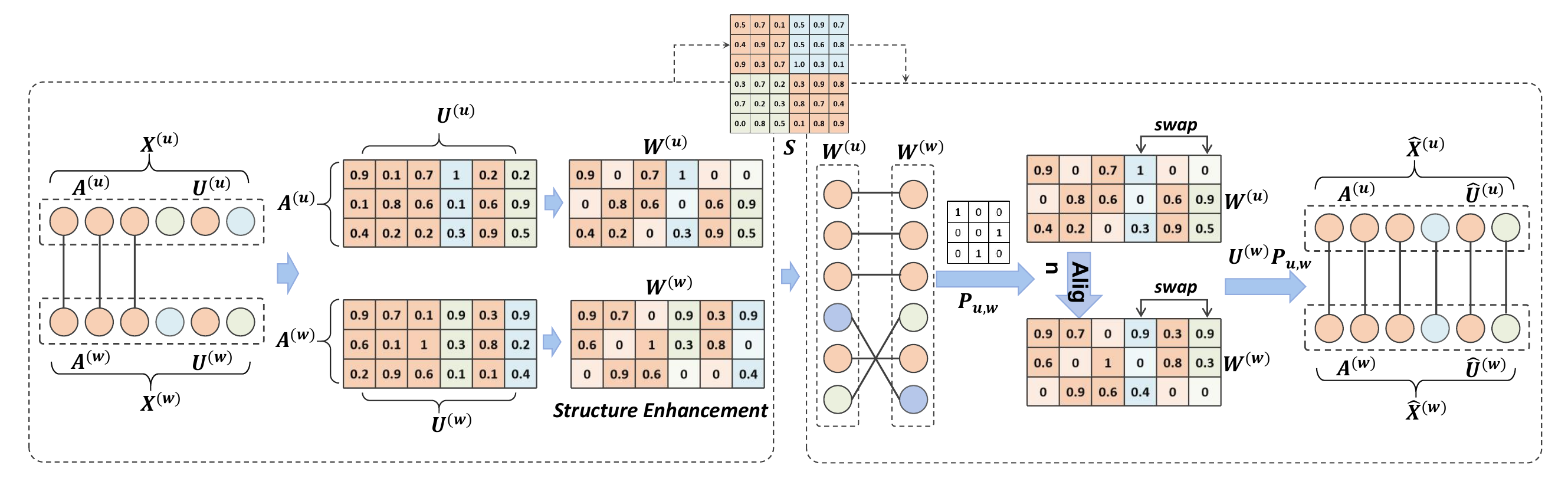}
\caption{The anchor graph structure alignment: The green and blue represent the misaligned samples. The upper part of the figure shows the anchor points of the samples and the structural filtering of the samples, while the lower part illustrates the preliminary alignment of the original samples guided by the joint embedding representation \textbf{W}.}\label{Fig:2}
\end{figure*}

\subsection{Graph Representation Learning}
Graph Representation Learning is a crucial methodology in machine learning, focusing on the extraction of valuable information from graph-structured data to derive low-dimensional, dense, and meaningful vector representations of nodes, edges, and entire graphs. These representations effectively capture the structural features and inter-node relationships within the graph, facilitating a range of downstream tasks such as clustering, community detection, and graph classification. Common graph construction techniques include K-Nearest Neighbors(KNN) and anchor graphs.
\subsubsection{K-nearest neighbor graph}
KNN \cite{guo2003knn,xie2024mgnr} graph construction is a widely used technique in graph learning and data mining. The core idea of this method is to identify the \textbf{K} nearest neighbors \cite{zhao2021k} for each data point, thereby defining the connectivity between data points. This means that the original dataset is transformed into a graph where nodes represent data points, and edges between nodes denote the proximity of data points.

Let $\mathbf{X}=\{x_1,\dots,x_n\}^T\in {R}^{n\times d}$ represent the data matrix, where \emph{n} is the number of data points, and \emph{d} is the dimensionality of the features. Each data point $x_i \in {R}^d$ belongs to one of \textbf{K} classes, 
$\mathbf{C}=\{c_1,\dots,c_k\}$. Given the entire dataset \textbf{X}, each data point is treated as a node in a graph, with each edge representing the affinity or relationship between pairs of nodes. If $x_i$ and $x_j$ have at least one in the other's \emph{k} nearest neighbours, they are considered connected. The edge weight between $x_i$ and $x_j$ is defined as,

\begin{equation}
    w_{i j}=\left\{\begin{array}{ll}
\exp \left(-\frac{\left\|x_{i}-x_{j}\right\|}{2 \sigma^{2}}\right), & \text { if } x_{i} \text { and } x_{j} \text { are connected } \\
0, & \text { otherwise }
\end{array}\right.
\end{equation}
where $\sigma$ is the bandwidth parameter.

The selection of the \textbf{K} value is crucial when constructing a graph structure based on the KNN algorithm. A small \textbf{K} value may fail to capture potential relationships between data points due to insufficient neighborhood samples, compromising the accuracy of the graph structure. Conversely, a large \textbf{K} value leads to redundant and complex graph structures, increasing computational costs, reducing sparsity, and ultimately impairing the efficiency and effectiveness of graph processing algorithms.
\subsubsection{Anchor graph}
The anchor graph is an effective technique for processing large-scale graph data, particularly in graph learning tasks such as spectral clustering, graph embedding, and semi-supervised learning. The core of this technique lies in selecting a set of representative points from the entire dataset, known as anchors, which serve as the basis for simplifying the data structure. By doing so, anchor graphs can capture the global structure of the data at a relatively low computational cost, significantly reducing the resources required to handle complex graph data and thereby improving the efficiency of large-scale graph learning tasks.  

Anchor graph \cite{wang2021fast,qin2024dual, Ma2026multi} techniques offer significant advantages in the field of graph-structured data processing, highlighted in several key areas. First, this method can efficiently construct graph structures with a relatively small number of edges, thereby reducing computational and storage costs while preserving the core structural information of the graph, making it suitable for large-scale data processing. 
Second, the construction method of anchor graphs is highly flexible. The flexibility arises from their ability to capture common features across different views through representative anchors, which is beneficial in addressing issues of incomplete and misaligned multi-view data, enabling effective information fusion in multi-view integration analysis. 
Lastly, the selection of representative anchors facilitates a more accurate revelation of the intrinsic relationships within the data and the central structure of the graph. This provides a solid foundation for subsequent graph analysis tasks, such as clustering, classification, and graph embedding \cite{cai2018comprehensive}, allowing anchor graph-based methods to achieve better performance and higher efficiency in these tasks.

\section{Method}
\label{}

Given a multi-view dataset \( \textbf{X} = \{\mathbf{X}^{(v)}\}_{v=1}^V \), where \( V \) is the number of views. \( \mathbf{X}^{(v)} = \left[x_1^{(v)}, x_2^{(v)}, \ldots, x_n^{(v)}\right] \in {R}^{n \times d_v} \), where \( n \) is the number of samples and \( d_v \) is the dimension of the original view representation. Due to separation in storage and processing, only a small portion \(\{\mathbf{X}^{(v)}\}_{v=1}^V\) is aligned. Thus, for two different views \( u \) and \( w \) (\( u \neq w \)), the original data can be reordered and divided into aligned data \( \mathbf{A}^{(u)}, \mathbf{A}^{(w)} \) with corresponding relationships and misaligned data \( \mathbf{U}^{(u)}, \mathbf{U}^{(w)} \) with no corresponding relationships. For the aligned data, \( \mathbf{A}^{(u)} = \left[a_1^{(u)}, a_2^{(u)}, \ldots, a_{n_a}^{(u)}\right] \) and \( \mathbf{A}^{(w)} = \left[a_1^{(w)}, a_2^{(w)}, \ldots, a_{n_a}^{(w)}\right] \), where \( a_i^{(u)} \) and \( a_i^{(w)} \) are from different views of the same sample. For the misaligned data, \( \mathbf{U}^{(u)} = \left[u_1^{(u)}, u_2^{(u)}, \ldots, u_{n_u}^{(u)}\right] \) and \( \mathbf{U}^{(w)} = \left[u_1^{(w)}, u_2^{(w)}, \ldots, u_{n_u}^{(w)}\right] \), where \( u_i^{(u)} \) and \( u_i^{(w)} \) are from different samples. Taking view \( u \) as an example, it is easy to derive a reordering matrix \( \mathbf{P}_{u,w} \in {R}^{n \times n} \) that satisfies \(\mathbf{X}^{(u)}=\mathbf{P}_{u,w}\left[\mathbf{A}^{(u)}, \mathbf{U}^{(u)}\right] \), where each row and each column of \( \mathbf{P}_{u,w} \) has exactly one entry equal to 1, and the rest are 0. In the paper, the term '$v$' view refers to all views, while '$u$' view and '$w$' view specifically denote two of these views.

\subsection{Anchor Graph Structure Alignment}
%% Inline mathematics is tagged between $ symbols.
Due to the sensitivity of KNN to the parameter $k$ in graph construction and its neglect of the global structure, we use anchor graphs to construct the connections between the nodes within each view. Specifically, for view $u$, an anchor set \( \mathbf{A}^{(u)} \) is defined, including all aligned nodes in the view. Similarly, view $w$ has a corresponding anchor set \( \mathbf{A}^{(w)} \). Importantly, since \( \mathbf{A}^{(u)} \) and \( \mathbf{A}^{(w)} \) provide direct contact points between different views, the anchor graph method can effectively synchronize information across different views through these anchors, achieving stable clustering and analysis results even in the presence of data inconsistencies. The structural graph of the Anchor-graph-based View Alignment is illustrated in Figure \ref{Fig:2}.

After the anchors are determined, a bipartite graph construction method is used to measure and compute the similarity between misaligned data \( \mathbf{U}^{(v)} \) and anchors \( \mathbf{A}^{(v)} \). The rows of the bipartite graph contain the misaligned data, while the columns contain the selected anchors as references. In this framework, the similarity between a misaligned sample \( u_i \) and an anchor \( a_j \) is defined as follows,

\begin{equation}
    \mathbf{W}_{ij}^{(v)}=
\frac{exp(-D^2(u_i^{(v)},a_j^{(v)})/\sigma^2)}{\sum_{j=1}^{n_u}\sum_{i=1}^{n_a}exp(-D^2(u_i^{(v)},a_j^{(v)})/\sigma^2)}  
\end{equation}
where $\mathbf{W}^{(v)}$ represents the re-representation of the misaligned data in the $v$th view; $\sigma$ denotes the standard deviation of the Gaussian similarity, which is typically set to 1, and $D^2(\cdot)$ represents the squared cosine distance. This method accurately identifies the anchor index set $\langle u_j\rangle^{v}$ most similar or closest to a given sample in a specific view 
$v$, establishing effective connections between samples and their nearest anchors, thus supporting analysis and processing. To strengthen the association between similar samples and reduce the influence of dissimilar samples, we apply a structural filtering process to the samples $\mathbf{W}^{(v)}_{ij}$ in $\mathbf{W}$, namely,
\begin{equation}
\label{W0}
    \mathbf{W}_{ij}^{(v)}=\left\{
\begin{array}
{cc}\mathbf{W}_{ij}^{(v)} & ,\forall i\in\langle u_j\rangle^v \\
0, & \text{otherwise}
\end{array}\right.
\end{equation}
By using structural filtering, the edge weights between dissimilar samples are set to zero, reducing the impact of irrelevant samples.

\begin{figure*}[t]
\centering
\includegraphics[width=0.99\textwidth]{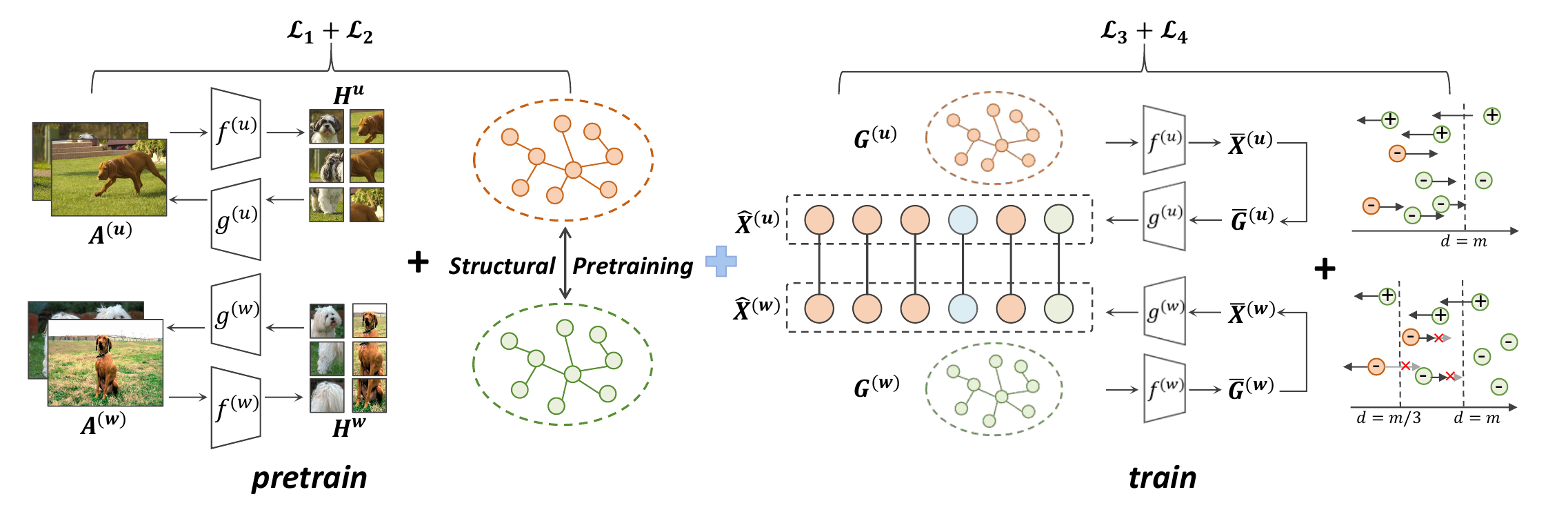}
\caption{Two-step training loss function: The loss in the pre-training stage consists of the sample reconstruction loss and the structural alignment loss, whereas the loss in the training stage consists of the sample reconstruction loss, structural reconstruction loss, and contrastive loss.}\label{Fig:3}
\end{figure*}

After calculating the similarity between the anchor set \( \mathbf{A}^{(v)} \) and the misaligned samples \( \mathbf{U}^{(v)} \), matrices \( \mathbf{W}^{(u)} \) and \( \mathbf{W}^{(w)} \) can be obtained. The re-representation of the misaligned matrices $\mathbf{W}^{(u)}$ and $\mathbf{W}^{(w)}$ links the global and local information of the graph, enabling the mapping of sample data from multiple views into a common dimensional space, thereby effectively leveraging the semantic information of samples from known classes. Subsequently, the two views 
$\mathbf{W}^{(u)}$ and $\mathbf{W}^{(w)}$ are aligned based on the similarity graph 
\( \mathbf{S} \). The similarity matrix \( \mathbf{S} \) is represented as follows, where \( {D}^2(\mathbf{W}_i^{(u)}, \mathbf{W}_j^{(w)}) \) also uses the cosine distance,

\begin{equation}
  \mathbf{S}^{(wu)}_{ij} = 
\begin{cases} 
\frac{\exp\left(-D^2(w_i^{(w)}, w_j^{(u)})/\sigma^2\right)}{\sum_{i \in \langle w_j \rangle^{u}} \exp\left(-D^2(w_i^{(w)}, w_j^{(u)})/\sigma^2\right)} & ,  i \in \langle w_j \rangle^{u} \\
0 &, \text{otherwise}
\end{cases}
\end{equation}
where $\langle w_j \rangle^{u}$ represents the index set of the $n$ nearest samples to $w_j^{(u)}$. Based on $\mathbf{S}^{(wu)}$, the best match for each column in matrix $\mathbf{W}^{(u)}$ is found in matrix $\mathbf{W}^{(w)}$ through the Hungarian algorithm, resulting in the permutation matrix $\mathbf{P}_{u,w}$, which reorders the misaligned samples and the overall samples in the multi-view data as follows,

\begin{equation}
\widehat{\mathbf{U}}^{(w)} = \mathbf{P}_{w,u} \mathbf{U}^{(w)}, \widehat{\mathbf{X}}^{(w)}=[\mathbf{A}^{(w)},\widehat{\mathbf{U}}^{(w)}]
\end{equation}

To obtain the aligned similarity matrix, we further reorder \( \mathbf{W}^{(w)} \) using the permutation matrix \( \mathbf{P}_{w,u} \),

\begin{equation}
\widehat{\mathbf{W}}^{(w)} = \mathbf{P}_{w,u} \mathbf{W}^{(w)}
\end{equation}

Since view $u$ is taken as the reference,

\begin{equation}
    \widehat{\mathbf{X}}^{(u)}=\mathbf{X}^{(u)}, \mathbf{X}^{(u)}=[\mathbf{A}^{(u)},\mathbf{U}^{(u)}]
\end{equation}

For the subsequent training of the graph convolutional network, the preliminarily aligned data in each view is used to construct a similarity graph \( \mathbf{G}^{(v)} \) using the KNN method.

\begin{equation}
    \left.\textbf{G}_{ij}^{(v)}=\left\{
\begin{array}
{cc}\frac{exp(-D^{2}(\widehat{x}_{i}^{(v)},\widehat{x}_{j}^{(v)})/\sigma^{2})}{\sum_{i\in(g_j)^{v}}exp(-D^{2}(\widehat{x}_{i}^{(v)},\widehat{x}_{j}^{(v)})/\sigma^{2})} & ,\forall i\in\langle g_j\rangle^{v} \\
0 & , \text{otherwise}
\end{array}\right.\right.
\end{equation}
Here $\langle g_j \rangle^{v}$ represents the index set of the $k$ nearest samples to $x_j^{(v)}$.

\subsection{Structure-Enhanced Feature Learning}

Structure-Enhanced Feature Learning consists of three components: structural filtering in Eq. \ref{W0}, which reduces the impact of irrelevant samples; the structural alignment loss 
$\mathcal{L}_2$ during pre-training, which guides structural learning; and the adjacency matrix in GCN learning, which guides feature learning along with the structural reconstruction loss 
$\mathcal{L}_3$.
To better capture the semantic features of the samples and the structural information between them, we pretrain the model with the training loss function, which includes the reconstruction loss $L_1$ and the structural alignment loss $L_2$, as shown below,
\begin{equation}
    \mathcal{L}_1=\left(\sum_{v=1}^V\sum_{k=1}^{n_a}\left(a_k^{(v)}-d_v\left(f_v(a_k^{(v)})\right)\right)^2\right)^{\frac{1}{2}}
\end{equation}

\begin{equation}
\mathcal{L}_2 = \left\| \bar{\mathbf{P}}_a - \mathbf{P}_a \right\|_F,\,
    \begin{cases}
f_u(\bar{\mathbf{A}}_b^{(u)})=f_u(\mathbf{A}_b^{(u)}) \\
f_w(\bar{\mathbf{A}}_b^{(w)})=\bar{\mathbf{P}}_a\cdot f_w(\mathbf{A}_b^{(w)})
\end{cases}
\end{equation}
Here, $f_v(\cdot)$ denotes the encoder of the $v$th view, and 
$d_v(\cdot)$ represents the decoder of the $v$th view. $P_a$ is the permutation matrix used to shuffle $\mathbf{A}_b^{(w)}$ during initialization, and 
$\bar{P}_a$ is the permutation matrix obtained through Hungarian alignment after learning the latent features of $\mathbf{A}_b^{(u)}$ and $\mathbf{A}_b^{(w)}$. 
$\mathbf{A}_b^{(v)}$ is extracted from a small portion of the aligned data $\mathbf{A}^{(v)}$.
Through the reconstruction loss and structural alignment loss, the model learns more accurate semantic representations, optimizing the relationship modeling between samples and better capturing the inherent structural features of the data.

After capturing the graph structure of multi-view data and completing pretraining and pre-alignment, a graph convolutional encoder is used to jointly learn samples and relational graphs to leverage graph relationships and capture the feature space of each view \cite{xia2021self}. In the GCN network, after processing through \( L \) layers of the graph convolutional encoder, to reconstruct the original information from these graph embeddings, decoder layers with the same number of layers as the encoder are employed. The purpose of these decoder layers is to correspondingly reverse the operations of the encoder layers, represented as follows,
\begin{equation}
\hat{\mathbf{H}}_{l-1}^{(v)} = \sigma \left( {\mathbf{D}}^{(v)-\frac{1}{2}} {\mathbf{G}}^{(v)'} {\mathbf{D}}^{(v)-\frac{1}{2}} \hat{\mathbf{H}}_l^{(v)} \hat{\mathbf{W}}^{(v)}_l \right)
\end{equation}

After passing through \( L \) layers of the decoder, the reconstructed node attribute matrix \( \hat{\mathbf{X}}^{(v)} = \hat{\mathbf{H}}_0^{(v)} \) is obtained. The reconstructed graph structure \( \bar{\mathbf{G}}_{ij}^{(v)} \) is represented by the pairwise similarity between the node embeddings \( h_i^{(v)} \) and \( h_j^{(v)} \) from the graph embedding \( \mathbf{H}^{(v)} \), calculated as follows,

\begin{equation}
\bar{\mathbf{G}}_{ij}^{(v)} = 
\begin{cases} 
\frac{\exp(-\mathcal{D}^2({h}_i^{(v)}, {h}_j^{(v)})/\sigma^2)}{\sum_{i \in \langle h_j \rangle^v} \exp(-\mathcal{D}^2({h}_i^{(v)}, {h}_j^{(v)})/\sigma^2)} &,\forall i \in \langle h_j \rangle^v \\
0 &, \text{otherwise}
\end{cases}
\end{equation}
where, $\langle h_j \rangle^v$ denotes the index set of the n nearest samples to $h_j^{(v)}$.
The loss function of the multi-view graph convolutional network is as follows,
\begin{equation}
\mathcal{L}_3 =  \sum_{v=1}^{V} \left\| \widehat{\mathbf{X}}^{(v)} - \bar{\mathbf{X}}^{(v)} \right\|_F^2 + \lambda_1 \sum_{v=1}^{V} \left\| \mathbf{G}^{(v)} - \bar{\mathbf{G}}^{(v)} \right\|_F^2 
\end{equation}
where, $\bar{\mathbf{X}}^{(v)}$ and $\bar{\mathbf{G}}^{(v)}$ represent the reconstructed sample matrix $\widehat{\mathbf{X}}^{(v)}$ and the reconstructed similarity graph $\mathbf{G}^{(v)}$, respectively, while 
$\lambda_1$ is the regularization parameter.

To further obtain accurate view correspondences, contrastive learning is introduced to guide the graph convolutional network in generating more discriminative features\cite{yang2021partially}. Node alignment between views resolves the positive and negative pair selection. Aligned nodes are considered positive pairs, while aligned and misaligned nodes form negative pairs. The distance between negative pairs is increased, while that between positive pairs is reduced,
\begin{equation}
\mathcal{L}_4=\frac{1}{2 N} \sum_{i=1}^N\left(Y \mathcal{L}_i^{p}+(1-Y) \mathcal{L}_i^{n}\right)
\end{equation}
Among them, \( N \) represents the number of contrastive pairs, \( Y = 1/0 \) marks positive/negative pairs, where \( \mathcal{L}_i^{\mathrm{pos}} \) represents the distance of positive pairs, and \( \mathcal{L}_i^{\mathrm{neg}} \) represents the distance of negative pairs, defined as follows,

\begin{equation}
    \mathcal{L}^{p}=d^2\left(h_i^{(u)}, h_j^{(w)}\right)+\max \left(m-d\left(h_i^{(u)}, h_j^{(w)}\right), 0\right)^2
\end{equation}

\begin{equation}
\mathcal{L}^{n}=\frac{1}{m} \max \left(m d^{\frac{1}{2}}\left(h_i^{(u)}, h_j^{(w)}\right)-d^{\frac{3}{2}}\left(h_i^{(u)}, h_j^{(w)}\right), 0\right)^2
\end{equation}
\( h_i \) is the graph embedding obtained through the graph convolutional network, \( d^2\left(h_i^{(u)}, h_j^{(w)}\right) \) is the distance metric function such as \( l_2 \) distance, and \( m \) is used to control the appropriate distance, calculated as follows,
\begin{equation}
m=\frac{1}{N_p} \sum d\left(h_i^{(u)}, h_i^{(w)}\right)+\frac{1}{N_n} \sum d\left(h_i^{(u)}, h_j^{(w)}\right)
\end{equation}

Subsequently, contrastive learning is introduced as a constraint in the feature learning process of the graph convolutional network, guiding the view alignment. By fully utilizing the aligned information, it increases the distinction between positive and negative samples, resulting in more accurate aligned features. 

Combining the graph convolutional network and contrastive learning, the final objective function of DAS-PMVC is as follows,
\begin{equation}
\mathcal{L} = \mathcal{L}_3 + \mathcal{L}_4
\end{equation}

\subsection{Dual Alignment Strategy}

The Dual Alignment Strategy includes anchor graph construction: as shown in the left part of Figure \ref{Fig:2}, where different views of the samples are mapped into the same-dimensional representation space through the construction of an anchor graph, resulting in sample joint embedding representations. The two-step alignment: as shown in the right part of Figure \ref{Fig:2}, the anchor graph alignment guides the first alignment of the samples, and Eq. \ref{Qwu} aligns the latent features $\mathbf{H}^{(v)}$ a second time through $\mathbf{Q}_{w,u}$.

After learning \( \mathbf{H}^{(v)} \) through the multi-view graph convolutional encoder, the Hungarian algorithm is used in the feature space to obtain the optimal graph matching of the bipartite graph \( \bar{\mathbf{G}} \) constructed from the feature space. This results in the alignment matrix \( \mathbf{Q}_{w,u} \) used to reorder the features \( \mathbf{H}^{(v)} \), represented as follows,

\begin{equation}
\label{Qwu}
\widehat{\mathbf{H}}^{(v)} = \mathbf{Q}_{w,u} \mathbf{H}^{(v)}
\end{equation}
Finally, spectral clustering is used on the aligned \(\widehat{\mathbf{H}}^{(v)}\) to obtain the clustering results.

\section{Experiments}\label{sec4}

\subsection{Experimental Setup}\label{ES}
Six multi-view datasets(Caltech20 \cite{lin2021completer}, BDGP \cite{wang2018partial}, Scene-15 \cite{fei2005bayesian}, Aloi \cite{schubert2010elki},  3Sources \cite{li2021incomplete}, BBCsports \cite{wen2018incomplete}) were used to validate the clustering performance of DAS-PMVC. Detailed information on the datasets is presented in Table \protect\ref{tab:dataset}. Meanwhile, eight advanced multi-view clustering algorithms were selected for comparison with DAS-PMVC, including PVC \cite{huang2020partially}, MvCLN \cite{yang2022robust}, EGPVC \cite{zhao2023end}, and ProImp \cite{li2023incomplete}, TCLPVC \cite{gao2025novel}, and EAGCP \cite{Ma2026multi}, which are leading partially multi-view algorithms.  Since AE2-Nets \cite{zhang2019ae2} and Cmib-Nets \cite{wan2021multi} do not handle partially aligned view data, an autoencoder was pre-trained to project the original data into the latent space, and the Hungarian algorithm was applied to establish their correspondence. 

\begin{table}
    \centering
    \caption{A description of the datasets.}
    \label{tab:dataset}
    \begin{tabular}{lccccc}
        \toprule
        Dataset & Samples & Views & Classes & Dimensions \\
        \midrule
        Caltech20 & 2386 & 2 & 20 & 48/40 \\
        BDGP & 2500 & 2 & 5 & 1750/79 \\
        Scene-15 & 4485 & 2 & 15 & 20/59 \\
        Aloi & 10000 & 2 & 100 & 77/64 \\
        3Sources & 169 & 2 & 6 & 3560/3631 \\
        BBCsports & 282 & 2 & 5 & 2582/2544 \\
        \bottomrule
    \end{tabular}
\end{table}

\begin{table*}
    \centering
    \caption{Comparison of clustering performance on Scene-15, BBCsports, 3Sources and Aloi datasets under 50\% view alignment rate.}
    \label{tab:clustering_performance1}
     \resizebox{\textwidth}{!}{
    \begin{tabular}{lcccccccccccc}
        \toprule
        \multicolumn{1}{c}{Datasets} & \multicolumn{3}{c}{Scene-15} & \multicolumn{3}{c}{BBCsports} & \multicolumn{3}{c}{3Sources} & \multicolumn{3}{c}{Aloi} \\
        \cmidrule(r){2-4} \cmidrule(r){5-7} \cmidrule(r){8-10} \cmidrule(r){11-13}
        Method & ACC & NMI & ARI & ACC & NMI & ARI & ACC & NMI & ARI & ACC & NMI & ARI \\
        \midrule
        AE2-Nets & 0.2625 & 0.2404 & 0.1203 & 0.3018 & 0.0235 & 0.0048 & 0.2998 & 0.0659 & 0.0022 & 0.0376 & 0.0996 & 0.0043 \\
        PVC & 0.3585 & 0.3943 & 0.1983 & 0.3582 & 0.0530 & 0.0214 & 0.3219 & 0.1146 & 0.0060 & 0.4235 & \underline{0.6716} & 0.3049\\
        Cmib-Nets & 0.1515 & 0.0813 & 0.0406 & 0.2649 & 0.0202 & 0.0040 & 0.2655 & 0.0522 & 0.0016 & 0.0604 & 0.1788 & 0.0101\\
        MvCLN & \underline{0.3817} & 0.3929 & \underline{0.2447} & 0.3957 & 0.0819 & 0.0609 & 0.3467 & 0.1239 & 0.0604 & 0.4440 & 0.6407 & 0.3431\\
        EGPVC & 0.3730 & 0.3882 & 0.1921 & 0.3759 & 0.1326 & 0.0435& 0.4201 & 0.1931 & 0.1547 & 0.4128 & 0.6588 & 0.2895 \\
        ProImp & 0.2267 & 0.2098 & 0.0955 & 0.3192 & 0.0483 & 0.0055& 0.3018 & 0.2736 & 0.0921 & 0.2126 & 0.4711 & 0.1193 \\
        TCLPVC & 0.3623 & \textbf{0.4010} & 0.2092 & 0.3440 & 0.0332 & 0.0314& 0.3254 & 0.0112 & 0.0048 & \underline{0.4988} & 0.7406 & \textbf{0.4176} \\
        EAGCP & 0.2841& 0.3031 &0.1358 & \underline{0.4468} & \textbf{0.2746} & \underline{0.1251}& \underline{0.4762} & \underline{0.3855} & \underline{0.1916} &0.2248 &0.5044 &0.1102 \\
        Ours & \textbf{0.3991} & \underline{0.3951} & \textbf{0.2468} & \textbf{0.4536} & \underline{0.2337} & \textbf{0.1723} & \textbf{0.5341} & \textbf{0.4010} & \textbf{0.2168} & \textbf{0.5045} & \textbf{0.6721} & \underline{0.3769}\\
        \bottomrule
    \end{tabular}}
    
\end{table*}

The view alignment rate of all datasets was set to 0.5, with the remaining samples randomly shuffled. ACC, NMI, and ARI were selected as the three metrics to evaluate the quality of the model's clustering results. Higher scores indicate better performance. 

\subsection{Result Analysis }\label{RA}

The experimental results are shown in Tables \protect\ref{tab:clustering_performance1} and \protect\ref{tab:clustering_performance3}. In the table, the best results are indicated in bold, and the second-best results are underlined. Experiments such as complexity analysis are provided in the appendix.

Experimental results show that the DAS-PMVC algorithm outperforms most comparative algorithms in clustering performance on multi-view datasets. Specifically, on the 3Sources dataset, the clustering ACC of DAS-PMVC exceeds that of the second-best algorithm by 5.79\%, and by 5.97\% on the Caltech20 dataset. Except for BDGP, DAS-PMVC achieves the best ACC on the remaining five datasets, demonstrating its exceptional clustering capability. This success is largely attributed to the algorithm's well-designed dual-alignment and structure enhancement strategies, which effectively leverage the semantic information from aligned data and structural information from different views.

\begin{table}
    \centering
    \caption{Comparison of clustering performance on BDGP and Caltech20 datasets under 50\% view alignment rate.}
    \label{tab:clustering_performance3}
    \setlength{\tabcolsep}{1mm}
    \resizebox{\linewidth}{!}{
    \begin{tabular}{lcccccc}
        \toprule
        \multicolumn{1}{c}{Datasets} & \multicolumn{3}{c}{BDGP} & \multicolumn{3}{c}{Caltech20} \\
        \cmidrule(r){2-4} \cmidrule(r){5-7}
        Method & ACC & NMI & ARI & ACC & NMI & ARI \\
        \midrule
        AE2-Nets & 0.4830 & 0.1603 & 0.1348 & 0.2157 & 0.0841 & 0.0390 \\
        PVC & 0.7740 & 0.5330 & 0.5129 & 0.4723 & \underline{0.6211} & 0.3502 \\
        Cmib-Nets & 0.5466 & 0.1921 & 0.1851 & 0.4734 & 0.6188 & 0.3814 \\
        MvCLN & 0.6514 & 0.4501 & 0.4382 & 0.4715 & 0.5500 & 0.3943 \\
        EGPVC & 0.7524 & 0.5695 & \underline{0.5126} & \underline{0.5134} & \textbf{0.6525} & 0.4163 \\
        ProImp & 0.4014 & 0.1232 & 0.1008 & 0.2362 & 0.2754 & 0.0981 \\
        TCLPVC & \textbf{0.9192} & \textbf{0.7787} & \textbf{0.8094} & 0.5080 &0.5750  & \textbf{0.5277} \\
        EAGCP & 0.3000 & 0.1363 & 0.0414 & 0.3210 &0.3573  & 0.1818 \\
        Ours & \underline{0.7796} & \underline{0.5812} & 0.4177 & \textbf{0.5731} & 0.5931 & \underline{0.4482} \\
        \bottomrule
    \end{tabular}}
    
\end{table}

On Aloi datasets, while DAS-PMVC maintains a leading position in clustering accuracy, its performance in NMI and ARI is somewhat lacking, not surpassing the comparative algorithms. This phenomenon might be due to the excessive number of categories in the Aloi datasets, which can cause an imbalance between categories and significantly affect NMI and ARI scores. This indicates that in cases with numerous categories, DAS-PMVC might have issues with clustering distribution balance.

The poor performance of DAS-PMVC on the BDGP dataset is attributed to the unsuitability of GCN for extracting latent features from the BDGP data. Both EAGCP and DAS-PMVC, based on GCN networks, perform poorly on BDGP. From a dataset perspective, BDGP contains visual and textual features extracted from Drosophila embryo images, which exhibit relatively weak and implicit local structural relationships. GCNs struggle to capture latent relationships in such data, as they inherently lack graph structure. Moreover, gene expression is similar within the same class of fruit flies, and the limited edges in the graph may exclude similar samples, hindering effective feature extraction.

Specifically, DAS-PMVC’s innovative dual anchor-based alignment strategy, combined with Structure-Enhanced Feature Learning and contrastive learning, ensures precise alignment of high-confidence nodes while demonstrating robustness in handling marg-
inally misaligned nodes. It evidences the significant advancements DAS-PMVC has made in enhancing clustering accuracy.
Overall, DAS-PMVC leads the comparative algorithms on average across three metrics on six datasets, especially achieving a breakthrough position in clustering accuracy. This demonstrates the effectiveness of our method on partially multi-view datasets and its advantages in multi-view data alignment.

\subsection{Ablation Experiment}\label{AE}

To thoroughly validate the effectiveness and superiority of our approach, a series of ablation experiments were designed and conducted. These experiments aimed to compare the performance differences between the anchor graph alignment strategy and traditional attribute-based alignment methods under different view alignment rates. In these experiments, particular attention was paid to the variation in alignment rate (ar), and the experiments were repeated ten times for each scenario, with the average taken to ensure the stability and reliability of the results. 
The corresponding experimental results are detailed in Table \protect\ref{tab:alignment_methods}.
\begin{table}
    \centering
    \caption{Comparison of two alignment methods on the Caltech20, 3source, and Scene-15 datasets under different view alignment rates.}
    \label{tab:alignment_methods}
    \setlength{\tabcolsep}{0.4mm}
    \resizebox{\linewidth}{!}{
    \begin{tabular}{lcccccc}
        \toprule
        \multicolumn{1}{c}{Datasets} & \multicolumn{2}{c}{Caltech20} & \multicolumn{2}{c}{3source} & \multicolumn{2}{c}{Scene-15} \\
        \cmidrule(r){2-3} \cmidrule(r){4-5} \cmidrule(r){6-7}
        ar/Method & Attribute & Anchor & Attribute & Anchor & Attribute & Anchor \\
        \midrule
        0.1 & 0.19 & \textbf{0.59} & 0.33 & \textbf{0.67} & 0.06 & \textbf{0.23} \\
        0.3 & 0.18 & \textbf{0.58} & 0.35 & \textbf{0.81} & 0.06 & \textbf{0.23} \\
        0.5 & 0.18 & \textbf{0.57} & 0.34 & \textbf{0.85} & 0.06 & \textbf{0.23} \\
        0.7 & 0.19 & \textbf{0.58} & 0.38 & \textbf{0.91} & 0.06 & \textbf{0.22} \\
        0.9 & 0.22 & \textbf{0.58} & 0.33 & \textbf{0.91} & 0.07 & \textbf{0.27} \\
        \bottomrule
    \end{tabular}}
    
\end{table}
\begin{figure}
\centering
\includegraphics[width=0.475\textwidth]{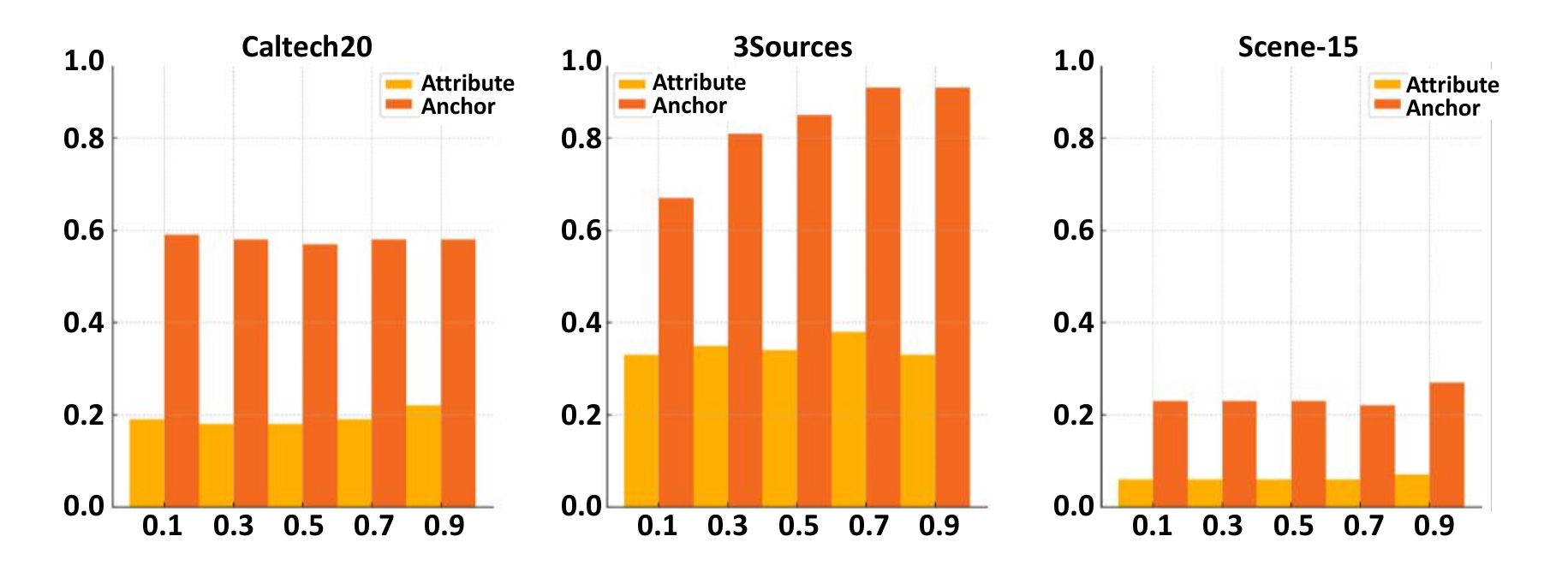}
\caption{Comparison of two alignment methods on the Caltech20, 3source, and Scene-15 datasets 
under different view alignment rates.}
\label{fig:exp}
\end{figure}

The experimental results indicate that DAS-PMVC significantly outperforms attribute-based direct alignment methods in view alignment performance, with the maximum improvement exceeding 50 percentage points.
Notably, increasing the proportion of initially aligned samples improves the alignment rate on 3Sources, while performance on Caltech20 and Scene-15 remains relatively stable across different alignment rates. This result not only highlights the unique advantages of our proposed method in handling view alignment issues but also demonstrates DAS-PMVC's strong adaptability and efficiency under various conditions.

Furthermore, to demonstrate the effectiveness of initial anchor-based view alignment for the clustering results of the model, the anchor graph alignment part was removed from DAS-PMVC, retaining only the graph convolutional contrastive learning network and the two-step alignment part, forming a comparative model called PMGCN. Experiments were conducted with both DAS-PMVC and PMGCN on the BDGP and Caltech20 datasets, with results shown in Table \protect\ref{tab:5}. The experimental results clearly indicate the impact of the anchor-based alignment module on improving DAS-PMVC's clustering accuracy, validating the effectiveness of the anchor-based view alignment method.

\begin{table}
    \centering
    \caption{Comparison of clustering results between DAS-PMVC and PMGCN on BDGP and Caltech20 datasets.}
    \label{tab:5}
    \setlength{\tabcolsep}{1mm}
    \resizebox{\linewidth}{!}{
    \begin{tabular}{lcccccc}
        \toprule
        \multicolumn{1}{c}{Datasets} & \multicolumn{3}{c}{BDGP} & \multicolumn{3}{c}{Caltech20} \\
        \cmidrule(r){2-4} \cmidrule(r){5-7}
        Method & ACC & NMI & ARI & ACC & NMI & ARI \\
        \midrule
        PMGCN & 0.5376 & 0.4655 & 0.3117 & 0.3502 & 0.3245 & 0.2637 \\
        DAS-PMVC & \textbf{0.7926} & \textbf{0.6084} & \textbf{0.4212} & \textbf{0.5682} & \textbf{0.5500} & \textbf{0.4209} \\
        \bottomrule
    \end{tabular}}
    
\end{table}
\section{Conclusion}\label{sec5}

This paper presents a framework for partial multi-view clustering via dual alignment and structure enhancement aimed at addressing the challenge of partial view alignment in multi-view clustering tasks. The dual-alignment strategy reduces errors from single-step alignment, while structure enhancement and structure-guided feature extraction mitigate the influence of irrelevant samples. Experimental results on multiple datasets demonstrate that DAS-PMVC significantly outperforms existing mainstream methods in handling partial view alignment. However, there is still the challenge of insufficiently utilizing structure for feature learning when the structural correlation between samples is weak. Future work will focus on exploring the applicability of GCNs.
%%
%% The acknowledgments section is defined using the "acks" environment
%% (and NOT an unnumbered section). This ensures the proper
%% identification of the section in the article metadata, and the
%% consistent spelling of the heading.
\begin{acks}
This work is supported by the National Natural Science Foundation of China (62572095), the Science and Technology Project of Liaoning Province (2024JH2/102600027), the Science and Technology Project of Dalian City (2024JJ12GX025, 2023JJ12SN029, and 2023JJ11CG005), and the Fundamental Research Funds for the Central Universities (DUT25YG246 and DUTZD25216).
\end{acks}

%%
%% The next two lines define the bibliography style to be used, and
%% the bibliography file.
\bibliographystyle{ACM-Reference-Format}
\bibliography{sample-base}

%%
%% If your work has an appendix, this is the place to put it.
\appendix

\end{document}